\documentclass[letterpaper,10pt,conference]{ieeeconf}
\IEEEoverridecommandlockouts
\usepackage{graphicx}
\usepackage{booktabs}
\usepackage{cuted}
\usepackage{capt-of}
\usepackage[hidelinks]{hyperref}
\hypersetup{pdftitle={Failure Modes of Vision-Language Models for Human Activity Understanding in Thermal HRI},pdfauthor={Han Hu and He Chen},pdfsubject={Preprint}}
\title{Failure Modes of Vision-Language Models for Human Activity Understanding in Thermal HRI}

\author{Han Hu and He Chen%
\thanks{Han Hu and He Chen are with the Department of Information Engineering,
The Chinese University of Hong Kong, Hong Kong SAR, China.
Email: {\tt\small hh022@ie.cuhk.edu.hk}, {\tt\small he.chen@ie.cuhk.edu.hk}.}}

\begin{document}
\maketitle
\thispagestyle{plain}
\pagestyle{plain}

\begin{abstract}
Thermal cameras enable robots to perceive human activities in darkness while revealing less personally identifiable appearance information and fewer details of the surrounding environment than RGB cameras. As vision-language models (VLMs), originally developed for RGB imagery, are increasingly integrated into embodied systems for scene understanding and high-level planning, thermal sensing offers a promising way to extend activity perception to low-light and privacy-sensitive settings. However, the failure modes of VLMs in thermal human activity understanding for human-robot interaction (HRI) remain underexplored. We fill this gap by evaluating seven locally deployable VLMs, including models developed for physical AI and robotics, on two person-centric thermal datasets. Our diagnostic tasks examine fine-grained action recognition, activity timeline reconstruction, and the ability to withhold unsupported answers. Paired RGB recordings serve as a performance reference, while controlled variations in frame sampling, thermal input presentation, and reasoning are used to assess whether the observed errors can be mitigated. Our results show that models can recognize broad activity categories but struggle with fine-grained actions such as coughing, sneezing, and yawning. They also produce inaccurate event timelines and hallucinate actions involving objects. Simple thermal input preprocessing and additional reasoning offer only limited, model-dependent gains. Abstention behavior also varies across models even under explicit instructions, limiting their reliability for practical robotic use.
\end{abstract}

\section{Introduction}
\label{sec:intro}

Vision language models (VLMs) increasingly form part of the
perception-and-reasoning stack of robotic systems, supporting visual
understanding, task decomposition, and high-level
planning~\cite{intelligence2026pi}. For robots that share
spaces with people, human activity is a central part of this
understanding: knowing what a person is doing, what they are
interacting with, and which events have just occurred provides
important context for deciding whether, when, and how to act. Activity
understanding therefore connects visual observation with downstream
reasoning and planning in embodied systems.

Robot perception of human activity in the home, however, is constrained by both lighting conditions and privacy concerns. Thermal video captures body shape and motion cues in darkness while revealing less facial and household appearance detail than RGB imagery~\cite{griffiths2018privacy,hu2026can}. This makes it well suited for human activity sensing in bedrooms and other private spaces, where a robot may need to detect a fall or recognize signs of discomfort even when the lights are off. When combined with the language interface of VLMs, thermal observations could support activity understanding at multiple levels of abstraction.

Prior work~\cite{hu2026can} found that VLMs recognize some thermal
activities but struggle with subtle gestures and movements, while more
recent thermal VLM studies target general thermal image understanding,
such as scene semantics and temperature
cues~\cite{cao2025irgpt,shrivastava2026thermeval,moshtaghi2025rgbthbench,thushara2026thermovl}.
Yet overall accuracy scores and isolated failure cases cannot
show where VLMs fail in understanding human activities from thermal
video, nor how much additional observation and reasoning can help. RGB video benchmarks also expose difficulties in fine-grained motion
recognition~\cite{hong2025motionbench,tu2026favor}, although the
different visual cues in thermal video limit direct transfer of these
findings. As a result, thermal video-based human activity understanding for HRI still lacks a concrete basis for choosing activity-understanding tasks, configuring
perception pipelines, and prioritizing improvements.

We address this gap through a diagnostic evaluation of seven locally
deployable VLMs on two person-centric thermal video datasets, testing
three aspects of activity understanding: distinctions between specific actions, recovery of action--object interactions and their temporal extent, and the decision between an action answer and backoff. Paired RGB--thermal recordings and controlled changes in observation and inference examine how failures respond to different conditions, and models developed for physical AI or robotics are compared with their general-purpose base-model reference.

Our systematic evaluation yields four key observations:

\begin{itemize}
    \item \textbf{Recurring action confusions.}
    Across seven models, five incorrect answer labels account for
    42.9\% of pooled exact-action errors. We also observe that recurring substitutions involve similar body movements or hand--object configurations, and these confusions from the test clips can cross functional categories, describing physiological events or potentially urgent behavior as routine activities, such as coughing labeled as touching the face and staggering as walking.

    \item \textbf{Activity understanding errors.}
    In long clips, models recover only 30.2\%--51.8\% of annotated
    events with the correct action and object labels, even without
    requiring accurate temporal boundaries. For harder joint tasks, their predictions also contain inaccurate temporal boundaries and hallucinated actions involving objects.

    \item \textbf{Limited intervention gains.}
    Across the two tested models, pseudocoloring, contrast enhancement, person cropping, denser frame sampling, and additional reasoning change exact-action accuracy by a limited $-5.9$\% to $+5.9$\% point relative to their respective paired baselines.

    \item \textbf{Unreliable abstention.}
    Models exhibit different answering tendencies but struggle to
    balance rejecting unsupported answers with retaining valid ones,
    even under explicit instructions. For example, Qwen3.6-27B correctly abstains on only 6.1\% of questions without a valid candidate, whereas RoboBrain2.5 answers only 3.2\% of questions containing one. We conclude this behavior as unacceptable for practical robotic application.
\end{itemize}

\section{Related Work}
\label{sec:related}

\paragraph{Human Activity Understanding for Robots}
Visual human activity understanding has been studied for robotic
assistance through action--object relationships and object
affordances~\cite{teo2012towards,koppula2013learning}.
Recent work also investigates fine-grained action recognition and
temporal segmentation in assembly, using pose cues, visual context,
and vision--language models~\cite{hyder2024action,zheng2024ducas,zheng2026compositional}.
In caregiving, OpenRoboCare~\cite{liang2025openrobocare} shows that
video-captioning models still omit fine-grained details and
misidentify actions.
These studies examine activity understanding in concrete robotic
settings, but their results with RGB or RGB-D observations leaves open how reliably the same action distinctions and temporal
structure can be recovered from thermal observations, where the available visual cues differ.

\paragraph{Thermal Human Sensing and Activity Understanding}
Thermal human sensing spans pedestrian detection, motion analysis,
and activity understanding. Benchmarks such as
LLVIP~\cite{jia2021llvip} support pedestrian detection under poor
illumination, establishing a foundation for locating people without
evaluating the activities they perform. Beyond detection,
InfAR~\cite{gao2016infar} investigates infrared action classification, while OrganicHAR~\cite{patidar2025organichar} discovers activity categories suited to privacy-preserving sensors, including thermal arrays. Thermal-IM~\cite{tang2023happened} exploits heat traces left by human--object contact to infer past human poses, motivated by household robotic assistance. Building on Thermal-IM, CREMA~\cite{yu2025crema} constructs ThermalQA to evaluate multimodal fusion for inferring past human actions. Recent work also evaluates VLMs on thermal activities. A zero-shot thermal recognition study~\cite{hu2026can} reports stronger recognition of large body movements than fine gestures and safety-related actions, while CUHK-X~\cite{jiang2026large} broadens multimodal evaluation to activity descriptions, contextual understanding, and action ordering. These findings establish both the potential of thermal activity perception and limitations that remain relevant to its use in robots.

Together, these studies highlight the importance of human activity understanding for robot perception and HRI, and explore thermal sensing as a promising source of information for these applications. Building on these studies, we take the evaluation of thermal activity perception for robots a step further. Beyond assessing whether models can recognize human activities under various environments, we design controlled experiments specialized for characterizing where errors concentrate, which actions are repeatedly confused, and which failure patterns are shared across models or specific to them.
\section{Evaluation Design}
\label{sec:design}

\begin{figure*}[t]
\centering
\vspace*{2pt}
\includegraphics[width=\textwidth]{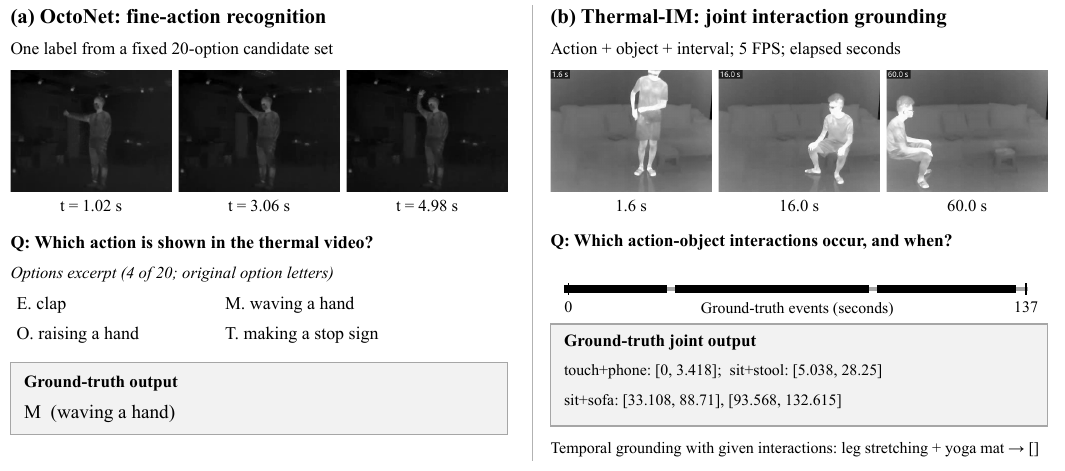}
\caption{Task examples: (a) fine-grained action recognition and
(b) action--object interaction temporal grounding.}
\label{fig:task_examples}
\end{figure*}

We structure the evaluation around three aspects of thermal activity
understanding: distinguishing specific actions, recovering
action--object interactions and their temporal extent, and choosing
between an action answer and backoff. OctoNet~\cite{yuan2026octonet} provides short
single-action clips for action recognition and candidate-set
interventions, while Thermal-IM~\cite{tang2023happened} provides recordings with annotated
interactions for recognition and temporal grounding.
Figure~\ref{fig:task_examples} provides examples of the thermal observations,
questions, and reference answers.

\subsection{Fine-grained Action Recognition}
\label{sec:tasks}

We manually select intervals from the OctoNet
recordings in which each participant repeatedly
performs the labeled action. After excluding one all-black clip, the
evaluation set contains 865 thermal clips at 320$\times$240
resolution, approximately 15 seconds each, covering 62 actions and
20 participants.

Each question asks the model to identify the observed action from
20 candidates and return its option letter
(Fig.~\ref{fig:task_examples}a). To probe distinctions between
similar actions, we select the 19 distractors using rules that
prioritize predefined groups of visually or semantically related
actions, then fill remaining slots from other labels using a fixed
seed. Candidate identities and order are shared across models, and
random-choice accuracy is 5\%. Exact-action accuracy measures
selection of the labeled action, while per-action accuracy identifies
which actions are repeatedly missed. To distinguish fine-label confusions from errors crossing broader
activity types, we map the same predictions and ground-truth labels
to 12 functional activity categories. We defined the mapping from the meanings of the action labels, grouping activities by everyday purpose or health and safety relevance, as motivated by the Index of ADL~\cite{katz1963studies} and OrganicHAR~\cite{patidar2025organichar}.
The same mapping was applied to ground-truth and predicted labels
across all models. For example,
walking and turning are grouped as locomotion, squats and jumping
rope as exercise, and hugs and handshakes as affiliative contact.
Category-wise accuracy is the fraction of predictions whose
functional categories match the ground truth. 

\subsection{Interaction Recognition and Temporal Grounding}
\label{sec:measures}

The official Thermal-IM test split contains
118 thermal recordings at 384$\times$288 resolution with 199
annotated action--object events. The recordings span 3.0--143.2
seconds (median 46.5), and 55 contain multiple annotated events.
Each event specifies an action, an object, and a start and end
time in seconds, kept at their original annotation precision.
We use synchronized RGB recordings for modality comparisons.

\paragraph{Joint task}
The model identifies every observed interaction from the
dataset-wide action and object vocabularies and returns a chronological list of \emph{(action, object, start time, end time)} entries (Fig.~\ref{fig:task_examples}b). This task evaluates both which interactions are recovered and what video segment they are located in the recording. 

\paragraph{Temporal grounding with given interactions}
To evaluate temporal grounding with known action and object identities, the model is given all distinct annotated action--object pairs in each recording, together with one absent pair. It is asked to return the start and end times of every occurrence of each
given pair, or an empty list if the pair is absent
(Fig.~\ref{fig:task_examples}b). Comparison with the joint task shows how temporal grounding changes when action and object identities are known.

\paragraph{Scoring}
For predicted and annotated intervals $I_p=[s_p,e_p]$ and
$I_g=[s_g,e_g]$, temporal intersection over union is
\[
\mathrm{tIoU}(I_p,I_g)
=
\frac{|I_p\cap I_g|}{|I_p\cup I_g|},
\]
using continuous interval lengths, $|I|=e-s$. Within each recording,
we match predictions and annotations with the same action--object
pair one-to-one, maximizing total tIoU.

Pooling counts across recordings, let $N_p$ and $N_g$ denote the
numbers of predicted and annotated events. Let $M_{\mathrm{pair}}$
be the number of matched pairs without a temporal-overlap threshold,
and $M_{0.5}$ the number of those matches with
$\mathrm{tIoU}\geq 0.5$. We report
\[
\mathrm{F1}_{\mathrm{pair}}
=
\frac{2M_{\mathrm{pair}}}{N_p+N_g},
\qquad
\mathrm{F1}_{\mathrm{event}}
=
\frac{2M_{0.5}}{N_p+N_g}.
\]
Pair F1 measures interaction recovery under the requested output
format; event F1 additionally requires sufficient temporal overlap.
Mean tIoU summarizes temporal agreement over matched events.
All three measures range from 0 to 1, with higher values
indicating better interaction recovery and temporal agreement.
Invalid temporal outputs count as empty predictions, contributing
missed events to both F1 scores.

\paragraph{Presence test for temporal interventions}
A separate 15-choice question asks which interaction is supported
by the video, offering 14 action--object pairs and a None option.
This test examines interaction selection without requiring
temporal boundaries. We measure the fraction of responses selecting the original target
interaction across complete-event clips, controls without an annotated
target event, and temporal-order transformations, keeping the question
and candidates unchanged.
The intervention conditions are described in
Section~\ref{sec:setup}.

\subsection{Candidate-set Backoff}

We test how the availability of the labeled action changes a
model's decision to answer or back off. This evaluation uses the same underlying principle as absent-answer detection in MM-UPD~\cite{miyai2025unsolvable}: testing whether models can withhold an answer when the correct candidate is unavailable. We use OctoNet rather than Thermal-IM to keep the backoff analysis directly comparable with the single-action recognition baseline, using identical clips and matched distractors across questions with and without the labeled action. 
For each OctoNet question, the \emph{ground truth (GT)-present} version retains the original 20 actions, while the \emph{GT-absent} version replaces the labeled action with another distractor, preserving the other 19. Both versions add the same backoff option, \emph{None of the listed actions / Cannot determine from the video}, for 21 choices in total. We evaluate these paired questions under the instruction settings described in Section~\ref{sec:setup}.

Correct-backoff rate is the fraction of GT-absent questions for
which the model selects backoff; false-backoff rate is the
corresponding fraction on GT-present questions. For GT-present
questions, coverage measures the fraction receiving an action
answer, and answered accuracy measures exact-action accuracy
among those answers. Together, these measures characterize both
the model's tendency to answer and the correctness of the action
labels it retains.
\section{Experimental Setup}
\label{sec:setup}

\subsection{Models and Inference}

We evaluate seven locally deployable VLMs without task-specific
adaptation to the evaluation data. The model set includes
Qwen3.5-4B~\cite{qwen35},
Qwen3.6-27B~\cite{qwen36},
Qwen3.8-27B~\cite{qwen38},
Gemma-4-31B-IT~\cite{team2026gemma},
Qwen3-VL-8B-Instruct~\cite{bai2025qwen3},
Cosmos3-Nano Reasoner~\cite{agarwal2026cosmos}, and
RoboBrain2.5-8B-NV~\cite{tan2026robobrain}.
Qwen3-VL serves as the general-purpose base-model reference
for Cosmos3 and RoboBrain2.5, allowing us to examine whether
additional training for physical AI or robotics improves thermal
activity understanding.

Since robotic use requires responses fast enough to inform ongoing
behavior, we adopt direct answers without extended reasoning as the
primary setting and evaluate additional reasoning separately.
All models are served locally with vLLM on a single NVIDIA RTX PRO 6000
Blackwell GPU; models in the 30B range run from NVFP4-quantized
checkpoints, while the smaller models run at native precision.
Models receive native video inputs sampled by their video processors
at a target rate of 5 FPS unless otherwise stated. For both the joint task and temporal grounding with given interactions, frames display their elapsed time in seconds, and both predicted intervals and annotations use continuous seconds (Fig.~\ref{fig:task_examples}b).

\subsection{Controlled Comparisons}

The frame-sampling, input-transformation, reasoning, and
temporal-evidence comparisons modify the thermal input or its
processing, so we run them with two representative models,
Qwen3.5-4B and Qwen3.6-27B. The RGB--thermal and explicit-backoff
comparisons leave the thermal input unchanged and cover all seven
models.

\paragraph{Frame sampling}
We examine whether denser observations help distinguish actions
and whether the benefit continues as more frames are provided.
We compare target rates of 2, 5, and
10 FPS on all 865 single-action clips, starting from the Qwen video
processor's default rate of 2 FPS~\cite{qwen35}.

\paragraph{Thermal input transformations}
Hand and face movements often occupy only a small part of a thermal
frame. We test whether changing the presentation of these cues
helps distinguish similar actions. JET pseudocolor and
percentile-based contrast enhancement emphasize intensity
differences; a tracked person crop enlarges the actor; and a
two-view input combines a tracked full-person crop with its upper-body crop to emphasize hand and face movements while retaining context.
We also estimate body keypoints from the thermal clips using
ViTPose++~\cite{xu2023vitpose++} and provide synchronized skeleton
videos alongside the thermal input.
After visually checking crop and skeleton quality and excluding
unreliable outputs, we evaluate 293 clips from five participants,
covering all 62 actions with three to five clips per action.
Each transformation is compared with its own matched baseline run.

\paragraph{Reasoning}
Reasoning modes allocate additional computation before producing
an answer, which often improves performance on general
tasks~\cite{team2026gemma}. We test whether this benefit
extends to interpreting thermal actions by comparing direct answers
with reasoning-enabled inference on 289 clips from the five-participant
subset used for input transformations, excluding four tap clips whose
labels do not precisely describe the recorded motion. We inspect the reasoning chains of 12 changed predictions to examine the visual interpretations accompanying corrected and newly introduced errors.

\paragraph{RGB--thermal comparison}
We compare synchronized RGB and thermal recordings in both the
joint task and temporal grounding with given interactions, using
the same inference settings. We report action--object recognition
and temporal grounding separately to characterize how the modality
gap differs between identifying interactions and locating them.

\paragraph{Temporal-evidence intervention}
We examine how event content and temporal order affect
interaction-presence answers. For each of the 118 Thermal-IM
recordings, we extract a complete annotated occurrence of the
target interaction, choosing the longest occurrence when it
repeats. Each event clip is paired with an equal-duration clip of the same
person performing other activities, drawn from a recording with no
annotated target action, while the question and candidates stay
unchanged. To perturb temporal order, we reverse the target event or cut it into consecutive blocks of at most three seconds each (at least
four blocks) and shuffle them. Shuffling preserves all frames and
within-block order with their original playback speed and same 5 FPS sample rate.

\paragraph{Backoff settings}
All seven models are evaluated on the 865 paired GT-present and
GT-absent questions with an explicit instruction to select backoff
when the model identifies that no listed action is supported or the visual evidence is insufficient. For Qwen3.5-4B, Qwen3.6-27B, Qwen3.8-27B, and Gemma-4-31B-IT, we also evaluate an option-only setting without
this instruction. Both settings retain the same backoff option
and candidate wording, allowing us to examine how explicit guidance
changes answering and backoff behavior beyond providing the backoff option alone.
\section{Results and Analysis}
\label{sec:results_analysis}

\begin{figure*}[t]
\centering
\vspace*{3pt}
\includegraphics[width=\textwidth]{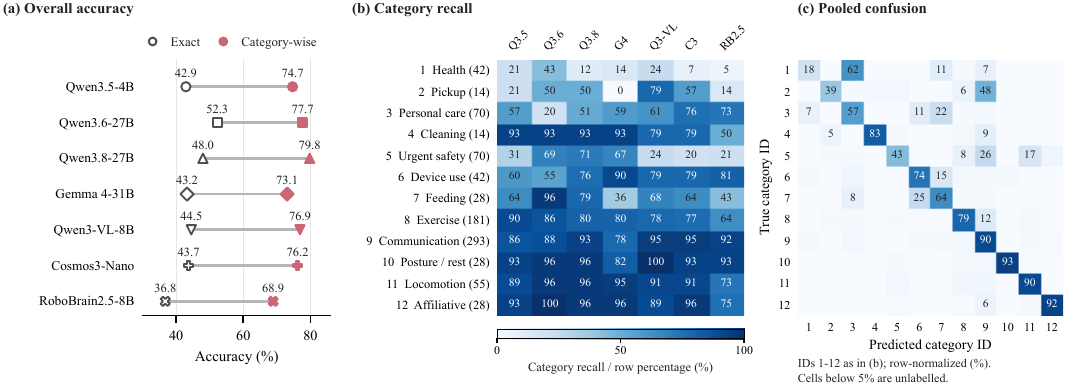}
\caption{Action recognition on 865 thermal clips. (a) Exact and
category-wise accuracy. (b) Per-category recall (clip counts in
parentheses). (c) Pooled category confusion, row-normalized.}
\label{fig:capability_granularity}
\end{figure*}

\subsection{Fine-grained Action Recognition}
\label{sec:action_recognition_results}

\paragraph{Confusions between specific actions}
The results reveal a gap between broad activity recognition and
fine-grained action identification (Fig.~\ref{fig:capability_granularity}a).
Large body movements are often recognized, while short hand-to-face
actions, physiological events, and gestures with similar trajectories
remain difficult. 

Across the seven models, the five most frequent incorrect answer labels
(waving, lifting a hand, touching the face, drinking, and turning)
account for 42.9\% of all
exact-action errors. Recurring substitutions include brushing teeth
predicted as drinking, yawning and coughing as touching the face, a stop
gesture as lifting a hand, and dodging as turning. These errors show
repeated confusion between actions sharing similar body movements or
hand--object configurations.

\paragraph{Consequences for activity meaning}
When mapped to functional categories, 50.8\%--61.1\% of each
model's exact-action errors retain the correct broad category.
But these similar movement-level confusions can also cross category boundaries (Fig.~\ref{fig:capability_granularity}b--c). In the tested clips, staggering is predicted as walking or turning, while pushing or punching someone is predicted as conversation. These substitutions present potentially urgent behavior as ordinary movement or communication, losing the distinction between safety-related and non-urgent activities in the resulting descriptions.

\paragraph{Sampling and thermal presentation}
Denser sampling and changes in thermal presentation give partial answers
to these confusions (Table~\ref{tab:input_comparisons}). Both models improve in exact-action accuracy when the target
sampling rate rises from 2 to 5 FPS, with little further gain at 10 FPS.
Person cropping provides small gains for both models, whereas JET and
contrast enhancement do not help. Skeleton
and upper-body views have model-dependent effects: adding skeletons
helps Qwen3.5-4B but hurts Qwen3.6-27B, while the full-plus-upper-body
input benefits Qwen3.6-27B but leaves Qwen3.5-4B unchanged. These
observation-side changes alleviate some errors without consistently
resolving fine-grained action confusions.

\begin{table}[t]
\caption{Frame sampling and input transformations.}
\label{tab:input_comparisons}
\centering
\small
\setlength{\tabcolsep}{4pt}
\begin{tabular}{lrr}
\toprule
Setting & Qwen3.5-4B & Qwen3.6-27B \\
\midrule
\multicolumn{3}{l}{\emph{Exact-action accuracy (\%), 865 clips}} \\
2 FPS & 37.8 & 46.9 \\
5 FPS & 42.9 & 52.3 \\
10 FPS & 43.7 & 52.7 \\
\midrule
\multicolumn{3}{l}{\emph{Accuracy change (pp), 293 clips}} \\
JET & $-2.0$ & $+0.3$ \\
Contrast & $+1.0$ & $+0.7$ \\
Person crop & $+2.7$ & $+2.7$ \\
Person crop + upper body$^{*}$ & $0.0$ & $+3.4$ \\
Thermal + skeleton & $+1.0$ & $-3.1$ \\
\bottomrule
\end{tabular}

\smallskip
{\footnotesize
$^{*}$ Compared with its matched full-person-crop baseline.\par}
\end{table}

\paragraph{Reasoning}
Additional reasoning does not resolve these confusions either. For
Qwen3.5-4B, 25 corrections are offset by 24 new errors, leaving accuracy
nearly unchanged. For Qwen3.6-27B, 30 new errors outweigh 13 corrections
and reduce accuracy by 5.9 percentage points. The inspected reasoning chains include misidentified objects, confused
body poses or motion patterns, and assumptions about unclear visual
details. Some repeatedly reconsider candidates or acknowledge ambiguity
before choosing a specific interpretation, including cases that overturn
a correct direct answer.

\subsection{Interaction Recognition and Temporal Grounding}
\label{sec:interaction_grounding_results}

\begin{figure*}[t]
\centering
\vspace*{3pt}
\includegraphics[width=\textwidth]{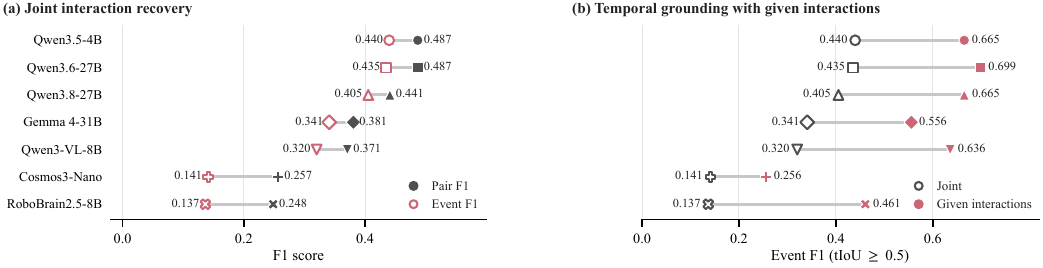}
\caption{Interaction recognition and temporal grounding on 118
thermal recordings. (a) Joint pair and event F1. (b) Event F1 for
the joint task and temporal grounding with given interactions.}
\label{fig:interaction_grounding}
\end{figure*}

\paragraph{Joint task}
In longer recordings, errors concern not only when events occur, but
which interactions are recovered. The joint predictions recover only part
of the annotated interactions (Fig.~\ref{fig:interaction_grounding}a).
Missed or misidentified events and invalid responses reduce pair F1, and
requiring temporal overlap further reduces event F1 because a correctly
named interaction can still receive inaccurate boundaries.

Differences in interaction recovery are not captured by isolated-action
accuracy alone. Cosmos3 and Qwen3-VL have similar exact-action accuracy
(43.7\% and 44.5\%), but their thermal pair F1 scores are 0.257 and
0.371, respectively; RoboBrain2.5 reaches 0.248. Thus, the additional
physical-AI or robotics specialization does not provide a consistent
advantage in recovering these interactions under direct answering.

\begin{figure}[t]
\centering
\includegraphics[width=\columnwidth]{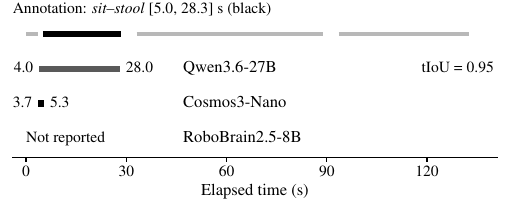}
\caption{The \emph{sit--stool} event in Fig.~\ref{fig:task_examples}b:
Qwen3.6 recovers it, Cosmos3 predicts a fragment, and RoboBrain2.5 omits it.}
\label{fig:grounding_example}
\end{figure}

Figure~\ref{fig:grounding_example} gives an example result of the four-event recording of Fig.~\ref{fig:task_examples}b. For \emph{sit--stool} at 5.0--28.3~s, Qwen3.6-27B returns 4.0--28.0~s ($\mathrm{tIoU}=0.95$),
Cosmos3 returns a 1.6-second fragment at the start, while RoboBrain2.5
reports one invented interaction \emph{leg stretching--stool} at 32--88~s. This example shows that same event can be either localized almost exactly or reduced to a fragment by
another, and replaced by an unrelated record from different models.

\paragraph{Temporal grounding with given interactions}
With action--object pairs given, the task evaluates their temporal localization. Event F1 rises in all seven
model comparisons once the interaction identities are given, with scores ranging from 0.256 to 0.699 (Fig.~\ref{fig:interaction_grounding}b).
We also observe that mean tIoU is higher in the joint task
(0.53--0.87 versus 0.48--0.77) when compared on a pair match. This gap is largely compositional rather than behavioral: on events recovered in both tasks, given identities yield equal or better boundaries for six of the seven models; the joint task simply skips the harder occurrences that the task with given interactions must attempt.

\begin{figure*}[t]
\centering
\vspace*{3pt}
\includegraphics[width=\textwidth]{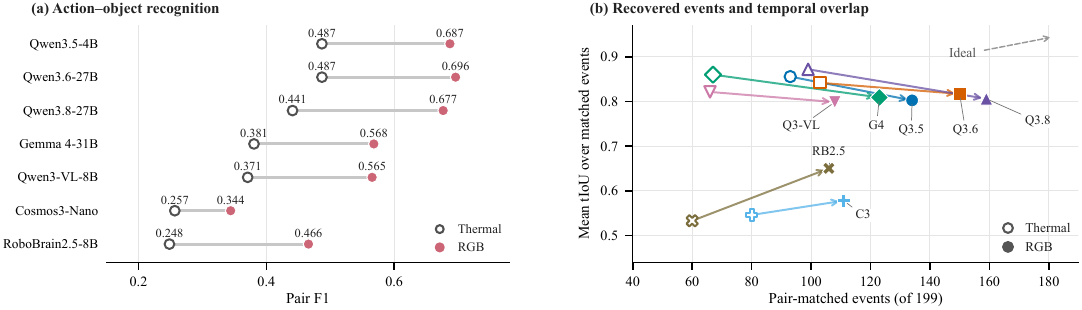}
\caption{RGB--thermal comparison in the joint task. (a) Pair F1.
(b) Thermal (open) to RGB (filled) operating points: pair-matched event
counts and mean tIoU over matched events. Scores use elapsed seconds.}
\label{fig:rgb_thermal_gap}
\end{figure*}

\paragraph{Paired RGB--thermal comparison}
The paired RGB comparison changes the visual input instead of the task information. RGB yields higher action--object pair F1 than thermal input across all seven models (Fig.~\ref{fig:rgb_thermal_gap}a). The mean
relative change in correctly matched event counts from thermal to RGB is
59.0\%. Qwen3-VL also outperforms Cosmos3 and RoboBrain2.5 in pair F1
under RGB input, indicating that their lower scores relative to the
base-model reference are not confined to thermal video.
Figure~\ref{fig:rgb_thermal_gap}b shows that RGB also recovers more
annotated events for every model, while mean tIoU over matched events
changes by at most 0.07 for all but Cosmos3 and RoboBrain2.5, for
which it improves.

\begin{figure}[t]
\centering
\includegraphics[width=\columnwidth]{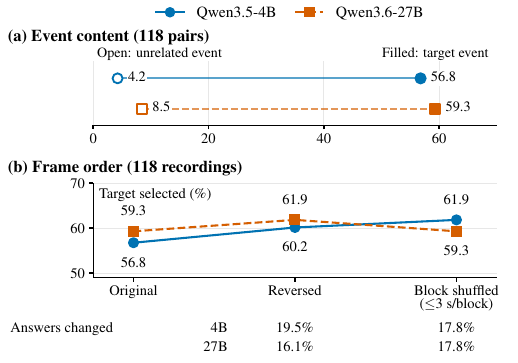}
\caption{Temporal-evidence interventions for tested models.
(a) Target selection with target events vs.\ unrelated activities.
(b) Target selection and answer changes after reversing or
block-shuffling events.}
\label{fig:temporal_evidence}
\end{figure}

\paragraph{Temporal-evidence intervention}
A complementary interaction-presence test examines whether recognition
errors remain when complete target events are shown and no boundaries
are requested. Selection of the target label drops by more than 50 percentage points for both models when the target event is replaced with an unrelated activity (Fig.~\ref{fig:temporal_evidence}a). This shows that their answers respond to the activity shown, but we still observe missed correct label for complete annotated target events, which means that relevant event content cannot fully resolve the interaction confusions.

Reversing events or shuffling blocks of at most three seconds changes 16.1–19.5\% of answers without a reliable decline in target selection. This pattern is understandable given the prevalence of sustained postures and object-contact interactions in the evaluated events. Posture and contact cues remain visible after reordering. Together with the sharp decline under content replacement, these results suggest that the models can exploit spatial evidence in thermal frames, while temporal order is less decisive for these interaction-presence judgments.

\subsection{Candidate-set Backoff}
\label{sec:backoff_reliability}

A further question is whether allowing models to withhold an
answer improves their performance. Backoff
exposes two opposite failures beyond action recognition
(Table~\ref{tab:backoff_summary}). Under explicit
instructions, Qwen3.6-27B, Qwen3-VL, and Cosmos3 continue answering
incorrectly on most GT-absent questions. At the other extreme,
RoboBrain2.5 rejects 98.5\% of GT-absent questions but answers only 28
of 865 GT-present questions; its 67.9\% answered accuracy therefore
accompanies only 3.2\% coverage. The same model reaches 36.8\%
exact-action accuracy without a backoff option
(Fig.~\ref{fig:capability_granularity}a): it is not simply unable to
classify the clips, yet it almost stops answering once a way out exists.

\begin{table}[t]
\caption{Explicit-backoff results (\%).}
\label{tab:backoff_summary}
\centering
\small
\setlength{\tabcolsep}{3pt}
\begin{tabular}{lrrr}
\toprule
Model & Coverage & Answered acc. & Correct backoff \\
\midrule
Qwen3.5-4B
& 50.1
& 55.0
& 67.7 \\

Qwen3.6-27B
& 97.2
& 52.1
& 6.1 \\

Qwen3.8-27B
& 85.9
& 49.1
& 25.0 \\

Gemma-4-31B-IT
& 92.7
& 45.0
& 16.1 \\

Qwen3-VL-8B & 93.5 & 46.2 & 15.3 \\
Cosmos3-Nano & 82.2 & 44.7 & 31.2 \\
RoboBrain2.5-8B & 3.2 & 67.9 & 98.5 \\
\bottomrule
\end{tabular}
\end{table}

\paragraph{Instruction comparison}
The paired instruction comparison examines whether explicit guidance prompt can adjust this tendency. When no explicit guidance is given, all four models rarely back off on GT-absent
questions, with correct backoff between 6.8\% and 13.8\%. With the
instruction added, Qwen3.5-4B rejects more GT-absent questions but also
backs off on roughly half of the GT-present questions; its answered accuracy
rises from 44.1\% to 55.0\% as coverage falls. Qwen3.8-27B and
Gemma-4-31B-IT exhibit milder versions of this trade-off. Qwen3.6-27B
instead becomes more willing to answer when the correct action is
absent. The same instruction therefore shifts models toward different
answering behaviors rather than uniformly improving backoff. These
responses make backoff difficult to rely on without evaluating the chosen
model: explicit guidance can suppress valid answers or fail to prevent
unsupported ones.

\section{Conclusion}
\label{sec:conclusion}

This paper examines where VLMs fail in understanding human activities
from thermal video by evaluating seven locally deployable models on two
person-centric datasets along action recognition, interaction
grounding, and candidate-set backoff. The failures concentrate in
three places. Models often recover the broad activity
but not the specific action: Qwen3.6-27B reaches 52.3\% overall
accuracy yet only 16.7\% on coughing, sneezing, and yawning, and five
labels account for 42.9\% of all action errors. Our evaluation result shows that these confusions can cross functional categories, presenting health-related or potentially urgent behavior as routine activities: coughing is labeled as touching the face, and staggering as walking or turning. Meanwhile, models recover only part of what happened and when: joint pair F1 stays between 0.248 and 0.487, and a correctly named event can still receive an unusable interval, leaving records that are plausible but wrong. And models cannot be relied on to decide whether to answer: with backoff
available, Qwen3.6-27B answers 97.2\% of GT-present questions but
correctly backs off on only 6.1\% of GT-absent questions, while
RoboBrain2.5 answers just 3.2\% of GT-present questions.

Even with complete event views, models continue to confuse interactions,
while the tested changes to frame sampling, thermal input presentation,
and reasoning provide only limited or inconsistent gains. RGB improves
action--object pair F1 across all seven models, and supplying interaction
identities improves event F1. These comparisons point to action and object
recognition as a major bottleneck. Reliable use in robots also requires
event labels and intervals supported by the video, together with backoff
that rejects unsupported interpretations while retaining valid answers.

\section{Limitations and Future Work}
\label{sec:limitations_future}

Three limitations bound these findings. First, the activity taxonomies
and recording protocols of OctoNet~\cite{yuan2026octonet} and Thermal-IM~\cite{tang2023happened} reflect their origins
in multimodal sensing and human-motion
analysis. Their fixed camera
setups and activity distributions differ from what a robot would
encounter in a given household, and our 5 FPS clip-based evaluation
includes neither active observation nor the coupling between activity
interpretation and robot behavior. Second, our analysis is behavioral:
we characterize how model outputs change across observation and
inference conditions, but not where inside a model the failures
arise, for example whether thermal cues are lost in visual encoding, or
whether temporal evidence is integrated across frames at all. Layer-wise
analyses~\cite{liu2026grounding} offer a complementary route to such
mechanistic questions. Third, our tasks measure recognition,
localization, and answering behavior, but they remain proxies for
robotic applications. We note that a thermal-video benchmark that
evaluates whether a robot responds appropriately to human activity is still missing.

Action understanding remains a hard engineering problem, and our
results show that thermal video makes it harder. Since no single change
we tested resolves these errors, we propose combining thermal-specific
adaptation, complementary sensing, and active observation.
Thermal-specific adaptation should target
the recurring action and object confusions identified here, from
specialized motion representations and recognizers~\cite{yokoyama2026plug}
to thermal-adapted encoders. Synchronized complementary sensors can
supply cues that thermal video resolves poorly, with fusion evaluated
by whether it corrects specific confusions rather than by aggregate
scores. A robot can also act: revisiting a suspected onset, moving
closer to an ambiguous object, or choosing a new viewpoint turns
observation into an active process. Agentic video-understanding
workflows such as PyVision-RL~\cite{zhao2026pyvision} extend this idea,
selecting relevant clips and checking uncertain interpretations against
stored observations; their value should be weighed against the
additional observation and computation they require.

\bibliographystyle{IEEEtran}
\bibliography{references}

\end{document}